\documentclass[letterpaper, 10 pt, conference]{ieeeconf}  

\IEEEoverridecommandlockouts                              

\usepackage{graphicx} 
\usepackage{booktabs}
\usepackage{multirow}
\usepackage{amsmath} 
\usepackage{bm}
\usepackage{algorithm}
\usepackage{algpseudocode}

\usepackage[table]{xcolor} 
\title{\LARGE \bf
Zero-shot Safety Prediction for Autonomous Robots \\with Foundation World Models
}

\author{Zhenjiang Mao$^{1}$, Siqi Dai$^{1}$, Yuang Geng$^{1}$, and Ivan Ruchkin$^{1}$%
\thanks{$^{1}$Zhenjiang Mao, Siqi Dai, Yuang Geng, and Ivan Ruchkin are with the Department of Electrical and Computer Engineering,
        University of Florida, Gainesville, FL, 32611, USA,
        {\tt\small \{z.mao, dais, yuang.geng, iruchkin\}@ufl.edu}}
}

\begin{document}

\maketitle
\thispagestyle{empty}
\pagestyle{empty}

\begin{abstract}
A world model creates a surrogate world to train a controller and predict safety violations by learning the internal dynamic model of systems. 
However, the existing world models rely solely on statistical learning of how observations change in response to actions, lacking precise quantification of how accurate the surrogate dynamics are, which poses a significant challenge in safety-critical systems.
To address this challenge, we propose foundation world models that embed observations into meaningful and interpretable latent representations. This enables the surrogate dynamics to directly predict interpretable future states by leveraging a training-free large language model. In two common benchmarks, this novel model outperforms standard world models in the safety prediction task and has a performance comparable to supervised learning despite not using any data. We evaluate its performance with a more specialized and system-relevant metric by comparing estimated states instead of aggregating observation-wide error.
\end{abstract}

\section{Introduction}

\begin{figure}[t!]
  \centering
  \includegraphics[width=0.75\linewidth]{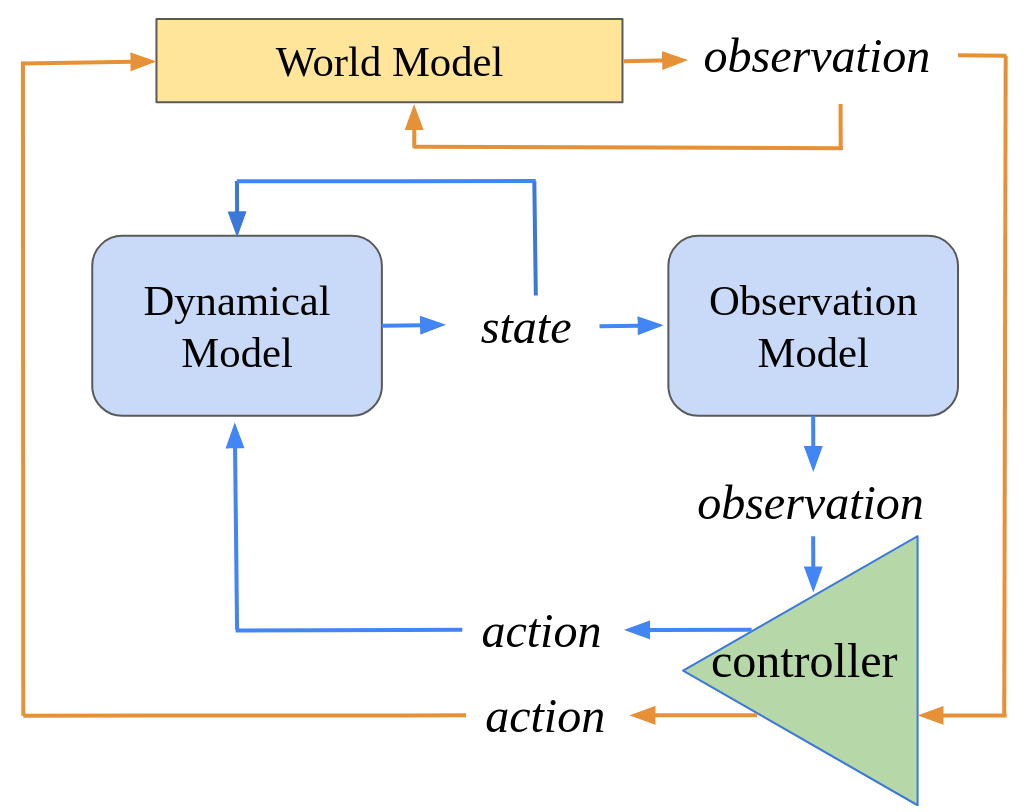}
  \vspace{-3mm}
  \caption{A dynamical model (blue flow) vs. a world model (orange flow).}
    \label{fig:dynamic-loop}
  \vspace{-6mm}
\end{figure}

A world model represents an understanding of how a robotic system works by learning how observations change with corresponding actions~\cite{ha2018worldmodels}. For example, it can describe how the image seen by a legged robot changes after taking several steps. Originally, world models were introduced to address data insufficiency when training reinforcement-learning controllers, which were limited by scarce real-world data~\cite{hafner2022mastering,micheli2023transformers}. This training typically requires the agent to operate in a real environment to gather extensive data, which makes it difficult and costly to explore diverse scenarios.
As shown in Fig.~\ref{fig:dynamic-loop}, a world model can learn the behavior of both the true dynamical model and observation models and thus become a surrogate of the real world.
 Some world models also learn rewards to support controller training. Not only do world models offer a new way for constructing better controllers by building a new generative \textit{world}, the surrogate dynamics also support quantifying the competence~\cite{acharya2022competency} and safety predictions~\cite{mao2024safe} for highly critical autonomous robots.

\begin{figure*}[t]
  \centering
  \includegraphics[width=\linewidth]{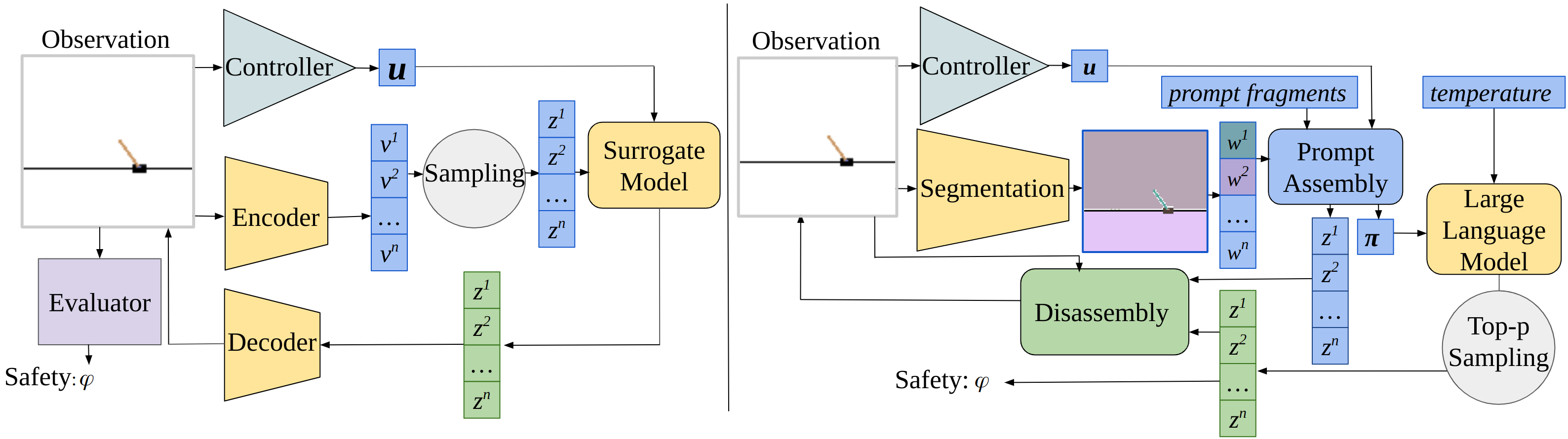}
  \caption{The structure of an existing world model (left) and the proposed foundation world model (right).}
    \label{fig:overview}
\end{figure*}

In the past, researchers primarily focused on training better controllers with world models, with insufficient attention given to how accurate the \textit{world} of world models is 
--- a crucial aspect for safety-critical robots. 
Typically, a world model does not directly predict observations, such as images from cameras or other sensor data, due to the constraints on time and computational resources. Instead, as shown on the left of Fig.~\ref{fig:overview}, it extracts useful features into latent representations with an encoder, which is part of a Variational Autoencoder (VAE)~\cite{kingma2022autoencoding}, and then predicts based on these representations. Usually, the performance of the world model is evaluated with the Mean Square Error (MSE) between the predicted observation and ground truth; this metric considers only the aggregate performance of the reconstruction --- and lacks the examination of the fine-grained and meaningful details of the prediction.
This coarse checking does not distinguish slight but critical differences (e.g., if a car's position shifts a few pixels, it may lead to a collision) that may cause devastating outcomes in safety predictions.
Also, the latent representations do not have a physical meaning in standard models and cannot be used to evaluate whether a predicted latent state is safe or has other important characteristics. As a result, it becomes necessary to develop an additional classifier to check such critical aspects, akin to a safety check as discussed in our earlier work~\cite{mao2024safe}, which requires more data, adds noise, and may suffer from distribution shift in predicted observations (compared to the real sensor data).

The rise of foundation models provides an opportunity to create meaningful representations with a zero-shot segmentation of observed images. Not only the interpretable representations can simplify the prediction and also the whole world model can also be implemented with a training-free architecture using Large Language Models, eliminating the need to collect and label training data.


This paper proposes \textit{foundation world models} that further reduce data requirements by using foundation models in two key elements of world models. First, to obtain interpretable latent representations, we use the \textit{Segment Anything Model} (SAM)~\cite{kirillov2023segment} to get the pixel positions of all objects in the observation. The important characteristics of latent states, in particular their safety (e.g., whether a collision has occurred or not), can be calculated based on these representations. Second, we predict the future position of these objects with a \textit{Large Language Model} (LLM). 
In addition, we introduce a more focused metric for the accuracy evaluation of predicted state. This metric helps us evaluate the quality of surrogate dynamics in a system-specific way. 

Our evaluation on two common simulated benchmarks demonstrates the value of foundation world models. Our foundation model not only demonstrates superior state prediction based on the newly proposed metric --- but also excels in safety predictions. We experiment with several baseline approaches, evaluation metrics, and different Large Language Models for latent prediction. 


In summary, this paper makes three contributions: 
\begin{enumerate}
    \item A training-free world model that combines foundation models with interpretable embedding and overcomes the distribution shift of the predicted observation, which occurs in standard world models.
    \item A segmentation-based metric for the accuracy of the surrogate dynamic prediction by quantifying the deviations of each object in the observation. 
    \item An experimental study of safety prediction where foundation world models show better performance despite not using any training data, compared to the existing world model and supervised learning methods.  
\end{enumerate}

Sec.~\ref{sec:prelim} introduces the details of world models and our problem description. Sec.~\ref{sec:approach} and Sec.~\ref{sec:exp} describe the foundation world models and the results of our experiments. In Sec.~\ref{sec:realated} and Sec.~\ref{sec:discussion}, we will review the related work and discuss the conclusion and future work.

\begin{figure*}[h]
  \centering
  \includegraphics[width=\linewidth]{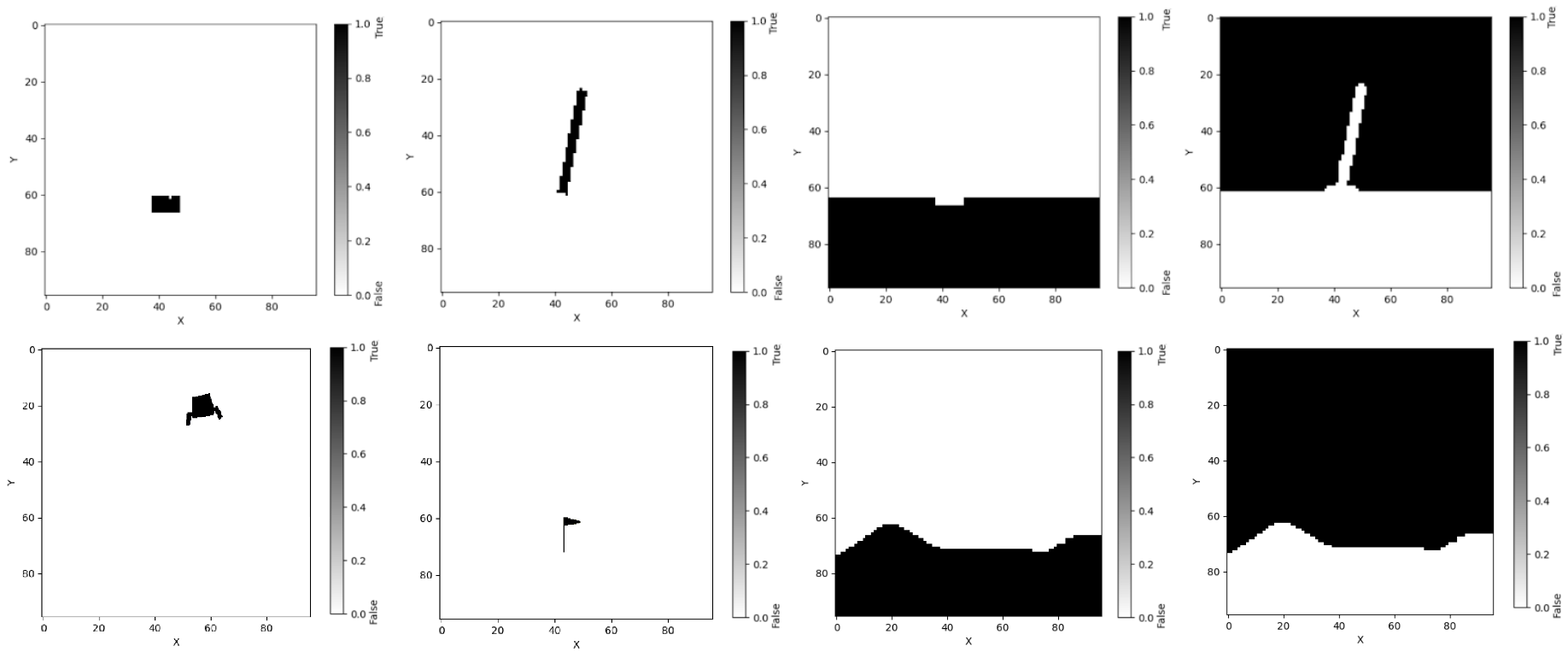}
  \caption{An example of segmentation matrices. Upper, from left to right: segmentation of the cart, pole, lower background, and upper background. Lower, from left to right: segmentation of the lander, lander point flag, lower background, and upper background.
  }
    \label{fig:seg}

\end{figure*}

\section{Preliminaries and Problem Statement}
\label{sec:prelim}

\subsection{Observation Prediction with World Models}

Consider a system $s$ with dynamical model ${f}$ in state $\bm{x}_{t}$, where action $\bm{u}_{t}$ is generated by an image-based controller $h$: $h(\bm{y}_{t}) = \bm{u}_{t}$. Each time step, the dynamics generates the state for the next time step: $f(\bm{x}_{t},\bm{u}_{t}) = \bm{x}_{t+1}$. The system also includes an observation model $g$ that converts the states $\bm{x}$ to observations $\bm{y}$: $g(\bm{x}_t) = \bm{y}_t$. 

The left side of Fig.~\ref{fig:overview} illustrates the architecture and operational workflow of the standard world model, composed of a decoder and an encoder~\cite{kingma2022autoencoding}. The \textit{encoder} ${\bm{E}}$ extracts useful features from images or sensor data ${\bm{y}} _{t} $ at time ${t}$ to build a distribution over latent vectors:  ${\bm{E}}({\bm{y}} _{t})={\bm{v}} _{t}=[\bm v^1_{t}, \bm v^2_{t},...,\bm v^n_{t}]$, where each $p^i$ is a one-dimensional Gaussian distribution: $N(\mu_t^i,\sigma_t^i)$. After sampling a latent vector $\bm{z}_t\sim \bm{v}_t$, the \emph{surrogate dynamical model} ${\bm{P}}$ (or simply surrogate model) predicts the future latent representation based on the past data: ${\bm{P}}({\bm{z}} _{t:t+m}, \bm{u}_{t:t+m}) = {\bm{z}} _{t+m+1} $. The \textit{decoder} ${\bm{D}} $ reconstructs a predicted observation $\hat{\bm{y}}$ from the latent representation: ${\bm{D}}({\bm{z}} _{t+1})=\hat{\bm{y}}_{t+1} $. The VAE and the surrogate model are trained sequentially. The optimization target of the VAE is to minimize the reconstruction error between the true image ${\bm{y}}_t$ and the reconstructed image $\hat{{\bm{y}}}_t$, as well as the KL divergence between the prior distribution and the latent distribution~\cite{kingma2022autoencoding}. In the surrogate model training, the mean squared error (MSE) is commonly employed as the loss function to minimize the error between the predicted image and the true image. In summary, the goal of a world model is to learn the following distribution:
$$p(\bm{y}_{t+m+1}\mid \bm{y}_{t:t+m};\bm{u}_{t:t+m}),$$
where $m$ is the input length of known observations. 
Thus, the world model $\bm{W}$ can be expressed as follows:

$$\bm{W}(\bm{y}_{t:t+m},\bm{u}_{t:t+m},\epsilon) = \hat{\bm{y}}_{t+m+1},$$
where $\epsilon$ is the source of randomness in the latent sampling. 

\looseness=-1
There is no single agreed-upon metric to evaluate the performance of a world model. In various applications of world models, the controller-training tasks are usually judged on how well the resulting controller performs compared with traditional methods~\cite{ha2018worldmodels,micheli2023transformers} in terms of reward. Traditionally, evaluation metrics like MSE are commonly adopted as the standard measurement criteria for quantifying the predictive capacity of world models. 
Accordingly, we compute the MSE of the world model's predicted observation in a pixel-wise manner: $$\text{MSE}({{\bm{y}}}, \hat{{\bm{y}}}) = \frac{1}{LW} \sum_{i=1}^{L} \sum_{j=1}^{W} (\hat{{\bm{y}}}^{(i,j)} - {\bm{y}}^{(i,j)})^2,$$ where $L$ and $W$ are the length and width of the image and ${\bm{y}}^{(i,j)}$ is the pixel value at the position $(i,j)$. 

We also consider another standard metric for visual prediction --- the \textit{Structural Similarity Index Measure} (SSIM): $$\text{SSIM}({{\bm{y}}}, \hat{{\bm{y}}}) = \frac{{(2 \mu_{\hat{{\bm{y}}}} \mu_{{{\bm{y}}}} + C_1)(2 \sigma_{{\bm{y}}\hat{{\bm{y}}}} + C_2)}}{{(\mu_{\bm{y}}^2 + \mu_{\hat{{\bm{y}}}}^2 + C_1)(\sigma_{\bm{y}}^2 + \sigma_{\hat{{\bm{y}}}}^2 + C_2)}},
$$ where $\mu$ and $\sigma^2$ are the mean and variance of an image over all pixels, $\sigma_{\bm{y}\hat{\bm{y}}}$ is the covariance between images, and $C_1$ and $C_2$ are stability constants used to avoid close-to-zero values in the denominator. This metric takes brightness, contrast, and structural differences into account. However, MSE and SSIM are aggregation metrics because they cover and summarize the whole image but ignore the more important objects in the images.
The next section will present our improvement over these metrics. 

\subsection{Object-based Prediction Metric}

\looseness=-1
Our insight, unused in earlier world models, is that the observation can be split into multiple meaningful objects by some segmentation algorithm $\Omega$: $\Omega(\textit{\textbf{y}})=[\bm{\omega}_1, \bm{\omega}_2, ...]$. Each $\bm{\omega}$ is a segmentation matrix with $True$ and $False$ pixels with the same size as  $\bm{y}$, as shown in Fig.~\ref{fig:seg} for two example systems. The $True$ elements reflect the position of the segmented object. These objects in the observation are semantically important elements of the world that are assumed to have a causal relation to future positions. The prediction of the objects' positions will evaluated separately, unlike e.g. the image-wise MSE that takes all pixels into account and is affected by inconsequential entities like static objects/background that occupy most of the observation frame. 


To tailor the evaluation to each object $\bm\omega$, we will calculate its \textit{centroid} $\bm\omega^{c} = (\bm\omega^{c}_x, \bm\omega^{c}_y)$ (i.e., its center of mass) as:

\begin{align*}
    \bm\omega^{c}_x &=\frac{1}{LW} \sum_{i=1}^{L} \sum_{j=1}^{W} j \cdot \bm{\omega}[i][j] \\ 
    \bm\omega^{c}_y &=\frac{1}{LW} \sum_{i=1}^{L} \sum_{j=1}^{W} i \cdot \bm{\omega}[i][j]
\end{align*}

Each prediction $\hat{\bm\omega}$ of a true object $\bm\omega$ will be quantified using the \textbf{centroid distance} (CD) defined with some norm $\mathcal{L}$, which can be for instance an L1 or L2-norm:
 $$\text{CD}(\bm\omega, \hat{\bm\omega}) = 
  \| \hat{\bm\omega}^{c} -\bm\omega^{c}\|_\mathcal{L} $$

For evaluating the image-level error between a prediction $\hat{\bm{y}}$ and a ground truth $\bm{y}$, we implement the segmentation algorithm to split the image into several parts: $\Omega({\bm y})=[\bm{\omega}_1, \bm{\omega}_2, ...,{\bm\omega}_n]$, $\Omega({\hat {\bm y}})=[\hat{\bm{\omega}}_1, \hat{\bm{\omega}}_2, ...,\hat{\bm\omega}_n]$. Then, we get a vector of CDs $[\text{CD}(\bm\omega_1, \hat{\bm\omega}_1),\text{CD}(\bm\omega_2, \hat{\bm\omega}_2),...,\text{CD}(\bm\omega_n, \hat{\bm\omega}_n)]$ that precisely reflects how well each object is predicted, rather than aggregating the errors of unrelated pixels. We leave the cases of missing and ghost objects for future work. 

 


\subsection{Safety Prediction Problem}

\looseness=-1
The \textit{safety predicate} $\varphi$ is a binary function of state $\bm x_t$ that determines the safety of the system at time $t$: $\varphi(\bm x_t)$. For example, it can indicate if the robot is dangerously close to an obstacle. The safety prediction problem~\cite{mao2024safe} is defined as follows: 
\begin{quote}
Given horizon $k>0$, a safety predicate $\varphi$, a sequence of $m$ observations and actions  from a known a controller $h$: [$\bm{y}_{t},\bm{y}_{t+1},...,\bm{y}_{t+m}$], [$\bm{u}_{t},\bm{u}_{t+1},...,\bm{u}_{t+m}$] from an unknown system $s$, determine whether the system satisfies the following logical formula for the future time moments: 
$$ \varphi(\bm{x}_{t+m}) \land \ldots \land \varphi(\bm{x}_{t+m+k}) $$

\end{quote}





Since safety prediction is binary, we use the F1 score to balance the evaluation of precision and recall. False-positive safety predictions are dangerous in safety-critical systems, so the false positive rate (FPR) is also taken into account.  

A significant challenge in standard world models is that the safety is not straightforward to infer from high-dimensional observations, so it necessitates an extra learning-based safety evaluator (often a Convolutional Neural Network, CNN) shown on the left of Fig.~\ref{fig:overview}. This CNN safety evaluator (as well as the controller) can suffer from distribution shift due to distorted predictions $\hat{\bm{y}}$ of standard world models originating in the decoder.  
Specifically, the conditional distribution of safety is the same for training and test datasets $p_{train}(\varphi(\bm x) \mid \bm y)=p_{test}(\varphi(\bm x) \mid \bm y)$, but the distribution of observations is shifted: $p_{train}(\bm y) \neq p_{test}(\bm y)$. This kind of distribution shift is covariate  shift~\cite{MORENOTORRES2012521}, which influences the accuracy of the CNN evaluator.

\begin{algorithm}[th]
\caption{Prompt assembly for state prediction}\label{alg:assemble}
\textbf{Input}: A series of observation $\bm{y}_{t:t+m}$, a segmentation model $\Omega$ and corresponding actions $(U=\bm{u}_{t},\bm{u}_{t+1},...,\bm{u}_{t+m})$, and prompt fragment $R$: instruction fragment $r_1$, input assembly fragment $r_2$, and prediction fragment $r_3$.
\\
\textbf{Output}: A full prompt $\pi$ and predicted states $\bm{z}$.
\\
\textsc{Function} \texttt{Assemble}($\bm{y}_{t:t+m},\Omega,U,R$):
\begin{algorithmic}[1]
\State $\bm{z} \gets [\ ]$ \Comment{Initialize an empty list}
\State $\pi \gets r_1$  \Comment{Add instructions at the beginning}
\For{$i$ from $t$ to $t+m$}
\State $\bm{\omega}_i\gets \Omega(\bm y_i)$
\For{$j$ from 1 to $\operatorname{len}$($\bm{\omega}_i$)}
\State $\pi \gets \pi+r_2$ \Comment{Add assembly prompt}
\State $\bm{z}_i^j \gets$$(\bm{\omega}_i^j)^c$ \Comment{Get the centroid}
\State $\pi \gets \pi+\bm{z}_i^j$  \Comment{Add the centroid}
\State $\pi \gets \pi+\bm{u}_M$  \Comment{Add action}
\State$\bm{z} \gets \bm{z}+\bm{z}_i^j$ 
\EndFor
\EndFor
\State $\pi \gets \pi+r_3$ 
\Comment{Add prediction order and specify the required output format  }
\State \textbf{return} $\pi,\bm{z}$

\end{algorithmic}
\end{algorithm}

\section{Foundation World Models}\label{sec:approach}

We propose \textit{foundation world models} that incorporate (a) a pre-trained foundation segmentation model into our proposed architecture as an encoder and (b) a large language model as a latent predictor. Specifically, we adopt the Segment Anything model (SAM)~\cite{kirillov2023segment} as the segmentation model to split the observation into several objects: $\Omega(\textit{\textbf{y}})=[\bm{\omega}_1, \bm{\omega}_2, ...]$. The extracted centroids $\bm\omega^{c}$ are used as latent representations $\bm{z}$, which are causally informative as future object positions are caused (in part) by past object positions. 
Combining the representations $\bm{z}$ with actions $\bm{u}$ and \textit{prompt fragments} $R = \{r_1, r_2, r_3\}$, we adopt the  Algorithm~\ref{alg:assemble} to form\footnote{We use ``+'' to denote string concatenation and adding to a state vector.} a complete prompt $\pi$. A Large Language Model takes prompt $\pi$ and returns a formatted prediction of ${\bm{P}}(\pi) = \hat{\bm{z}}$.  Three types of pre-prompts in $R$ are used to assemble a full prompt: the first $r_1$ is to describe the whole task, e.g., \textit{``Suppose I have a sequence of actions and states of a system, please predict the next step's state given the following information.''} Next, there's a loop to fuse the state description $r_2$ and the corresponding values \textit{``The time, states and the action for this step are:'}. We will repeat to add $r_2$ for m times to add all the input information. Finally, we add the state prediction instruction $r_3$: \textit{``Can you predict the state for the next moment? Please only give me your prediction values as a list''} at the end of the prompt to form the complete $\pi$.

To implement LLM-based prediction of latent states, we choose two models: GPT 3.5 provided by OpenAI and Gemma 7B-it based on the Gemini technology~\cite{geminiteam2023gemini}. 
One challenge is that not all segmented objects prove to be useful; for instance, objects like the white background in the cart pole system are uninformative by themselves. These non-essential objects require more tokens in the inference of the Large Language Model, which increases the computational cost and also potentially reduces prediction accuracy.
We eliminate this redundant information before the step of prompt assembly by removing objects whose centroids do not change at any time in the past observations. Future work can explore other causal techniques~\cite{yang2023causalvae} to focus on the prediction-relevant objects. 

\begin{algorithm}
\caption{Output disassembly into observation}\label{alg:disassemble}
\textbf{Input}: Predicted state $\bm{z}_{t+m+1}$, current state  $\bm{z}_{t+m}$, current observation  $\bm{y}_{t+m} $, and segmentation algorithm  $\Omega$.
\\
\textbf{Output}: Predicted observation $\bm{y}_{t+m+1}$
\\
\textsc{Function} 
\\
\texttt{Disassemble} ($\bm{z}_{t+m+1},\bm{y}_{t+m},\bm{z}_{t+m},\Omega$):
\begin{algorithmic}[1]
\State $\bm{y}_{t+m+1}\gets \bm{y}_{t+m}$  \Comment{Copy most recent observation}
\For{$i$ from 1 to $\bm{len}$($\bm{z}_{t+m+1}$)}
\State $\bm{\omega}_i\gets \Omega(\bm y_i)$
\State $\bm{\delta}^i \gets \bm{z}_{t+m+1}^i-\bm{z}_{t+m}^i$  \Comment{Compute the displacement of each object} 
\For{$pixel=True$ in $\bm{\omega}_i$}
\State $px \gets pixel.x$ \Comment{x position of the pixel} 
\State $py \gets pixel.y$ \Comment{y position of the pixel}
\State SWAP($\bm{y}_{t+m+1}[px][py]$, $\bm{y}_{t+m+1}[px+{\delta}^N_x][py+{\delta}^N_y]$)  \Comment{Move the object into the predicted position}
\EndFor
\EndFor
\State \textbf{return} $\bm{y}_{t+m+1}$
\end{algorithmic}
\end{algorithm}

\looseness=-1
World models perform stochastic prediction to ensure diverse outputs. In existing world models, this randomness is implemented by sampling latent states in the VAE, the diversity of which can be controlled with the prior. In our LLM-based world model, we implement randomness with \textit{top-p sampling}~\cite{holtzman2020curious}, for which temperature and threshold are two hyperparameters to control the diversity of the outputs. The top-$p$ sampling restricts the outputs to combinations of words with a cumulative probability higher than the threshold $p$.

\begin{table*}[th!]
\centering
\begin{tabular}{llllllll|llllll}
        \toprule

\multirow{3}{*}{} & \multirow{3}{*}{} & \multicolumn{6}{c}{Horizontal position CD error} & \multicolumn{6}{c}{Vertical position CD error}  \\

        \midrule
     Method       &  Input length      & k=10     & k=20        & k=30     & k=40        & k=50     & k=60   & k=10     & k=20        & k=30     & k=40        & k=50     & k=60     \\ \hline
 \rowcolor{gray!20}
VAE \& MDN-LSTM &        & 7.217 & 7.096 & 7.030 & 7.138 & 7.046 & 7.125 & 3.666 & 3.417 & 3.280 & 3.131 & 3.058 & 2.908

  \\

 \rowcolor{gray!20}
SAM \& MLP  &  & \textbf{.0910} & \textbf{.1698} & \textbf{.2670} & \textbf{.3840} & \textbf{.5017} & \textbf{.5975} & .4230 & .7957 & \textbf{.1092} & \textbf{.3905} & .6713 & .8780

\\
 \rowcolor{gray!20}
SAM \& LSTM  &$m$=1  & .1305 & .2558 & .3594 & .4983 & .6096 & .7356 & .2586 & \textbf{.5135} & .7338 & .9390 & \textbf{.1772} & \textbf{.3371}

\\
SAM \& GPT 3.5   &             & 3.122 & 1.175 & 5.985 & 3.738 & 4.793 & 3.905 & 2.969 & 1.195 & 2.147 & 4.572 & 4.794 & 6.292  \\
SAM \& Gemma   &        & .4955 & .3124 & .4827 & .7379 & .7316& .7568 & \textbf{.2385} & .6611 & 8.739 & 5.199 & 5.191  &5.736
\\
 \hline
  \rowcolor{gray!20}

VAE \& MDN-LSTM &           & 
 3.666 & 3.417 & 3.280 & 3.131 & 3.058 & 2.908 & 3.605 & 3.359 & 3.277 & 3.087 & 3.057 & 2.884 

\\
 \rowcolor{gray!20}

SAM \& MLP  &  & \textbf{.0799} & \textbf{.1444} & \textbf{.2075} & \textbf{.2689} & \textbf{.3363} & \textbf{.4206} & \textbf{.3107} & \textbf{.5970} & \textbf{.8939} & \textbf{.1624} & \textbf{.4233} & \textbf{.7062}

\\
 \rowcolor{gray!20}

SAM \& LSTM  & $m$=2 &.1262 & .2461 & .3435 & .4253 & .5065 & .5829 & .3536 & .6680 & .9888 & .2648 & .5389 & .8033

\\
SAM \& GPT 3.5   &            & .1495 & .2244 & 1.088 & 1.057 & 3.014 & 2.730 & 1.853 & 4.176 & 2.256 & 3.359 & 3.401 & 4.404 
\\
SAM \& Gemma     &    &    .6474 & 6.449 & 3.146 & .3081 & 5.564& 3.261& .8410 & 7.952 & 3.193 & .4780 & 4.426 & 8.634
   \\
 \hline

 \rowcolor{gray!20}

VAE \& MDN-LSTM &        & 
3.605 & 3.359 & 3.277 & 3.087 & 3.057 & 2.884 & 3.612 & 3.323 & 3.327 & 3.116 & 2.894 & 2.826

 \\
 \rowcolor{gray!20}

SAM \& MLP  &  & \textbf{.0684} & \textbf{.1333} & \textbf{.2039} & \textbf{.2795} & \textbf{.3313} & .4124 & \textbf{.2481} & \textbf{.4197} & \textbf{.5750} & .6773 & .7743 & .8921

\\
 \rowcolor{gray!20}

SAM \& LSTM  & $m$=4 & 
.1396 & .2654 & .3928 & .5002 & .5980 & .6748 & .3145 & .5833 & .9137 & \textbf{.1572} & .3273 & \textbf{.5452} 

\\
SAM \& GPT 3.5   &            & .1011 & .1654 & 1.438 & 1.012 & 3.704 & 2.924 & 1.167 & .9396 & 5.967 & 2.723 & 2.396 & 6.000 
 \\
SAM \& Gemma    &        & .2000 & .5749 & .5357 & .4636 & .8839 & \textbf{.1608} & .6410 & .9549 & 1.9403 & .1722 & \textbf{.2840} & .9854 
 \\
 \hline

  \rowcolor{gray!20}

VAE \& MDN-LSTM &           &  3.612 & 3.323 & 3.327 & 3.116 & 2.894 & 2.826 & 3.622 & 3.390 & 3.216 & 2.973 & 2.880 & 2.866

 \\

 \rowcolor{gray!20}

SAM \& MLP  && 4.938 & 7.279 & 9.578 & 27.14 & 45.94 & 29.42 & 2.290 & 75.62 & 51.35 & 30.91 & 54.48 & 93.20

\\
 \rowcolor{gray!20}

SAM \& LSTM  &$m$=8  & .1892 & .3761 & \textbf{.5760} & .7560 & .9665 & \textbf{.2009} & \textbf{.3077} & .5708 & .8250 & .9926 & \textbf{.1637} & .2818

\\
SAM \& GPT 3.5  &             &\textbf{.1001} & \textbf{.1374} & 2.412 & 1.180 & 4.247 & 5.231 & .3391 & .6566 & 3.298 & 2.180 & 3.143 & 7.534 
 \\

SAM \& Gemma      &        &  .8003 & .1471 & .9705 & \textbf{.2189} & \textbf{.5253} & .5808 & .3354 & \textbf{.5219} & \textbf{.2950} & \textbf{.1237} & .9888 & \textbf{.0901}
  \\

        \bottomrule

\end{tabular}

\caption{State prediction performance for the lunar lander: the horizontal and vertical centroid distance (CD) errors. Gray rows indicate the use of additional data. The range of position is [0, 20].}
\label{tab:cdlunar}
\end{table*}

\begin{table*}[th!]
\centering
\begin{tabular}{lllll|lll|lll|lll}
        \toprule

\multirow{3}{*}{Method} & \multirow{3}{*}{Input length} & \multicolumn{6}{c}{Position CD error (pixel range: [-48, 48])} & \multicolumn{6}{c}{Angle (degree range [-180\textdegree , 180\textdegree ])}\\ & & \multicolumn{3}{c}{Upright} & \multicolumn{3}{c}{Falling }  & \multicolumn{3}{c}{Upright} & \multicolumn{3}{c}{Falling } \\

        \midrule
            &        & k=10     & k=20        & k=30     & k=10        & k=20     & k=30   & k=10     & k=20        & k=30     & k=10        & k=20     & k=30     \\ \hline
 \rowcolor{gray!20}
VAE \& MDN-LSTM &  \multirow{5}{*}      & 3.767       & 3.628       & 4.807       & 9.083      & 9.046    &11.93 &   25.48  &24.16&25.88&36.76&37.26&37.72  \\

 \rowcolor{gray!20}
SAM \& MLP  &  & 1.301  & 2.750  & 4.304  & \textbf{3.248}  &  5.501 & 9.101  & 4.638  & 8.784  & 14.54  &16.05 & 21.89 & 29.40

\\
 \rowcolor{gray!20}
SAM \& LSTM  & $m$=1 & \textbf{1.276}  & \textbf{2.541} & 4.141  & 4.391 & \textbf{4.891}  & \textbf{3.976 } & \textbf{4.557} &\textbf{7.836} & 9.555 & 19.31  &\textbf{20.38}  &\textbf{8.489}

\\
SAM \& GPT 3.5   &             & 1.983       & 3.008      & \textbf{3.604}       &   5.401    & 8.537   &10.02 &6.187 &8.270&\textbf{9.022}&20.97 &30.18 &31.47 \\
SAM \& Gemma   &        & 1.935       & 3.993 & 4.582       & 3.745   & 8.411    & 10.236& 4.882       & 10.63        & 12.58 &  \textbf{8.110}     & 34.66       & 36.69 \\
 \hline
  \rowcolor{gray!20}

VAE \& MDN-LSTM &           & 3.627      & 3.932       & {4.747}       & 9.052     & 9.045    & 11.87 &26.29 &25.43&24.50&37.44 &36.39 &36.77 \\
 \rowcolor{gray!20}

SAM \& MLP  &  & 2.264  & 3.767 & 6.527  & 3.662 & \textbf{4.074}  &  \textbf{7.453}  & 8.276  & 17.64  & 36.89  & 19.16  & 27.03  & 50.79

\\
 \rowcolor{gray!20}

SAM \& LSTM  & $m$=2 & 2.571  & 4.815  & 6.490  &5.777  & 8.626  & 9.332  & 8.226 & 16.61  & 29.53  & 22.51  & 33.27 &45.33

\\
SAM \& GPT 3.5   &            & \textbf{1.685 }      & 3.704       & 5.778       &  \textbf{2.790 }     & 5.455    &10.40 & 6.330  &13.27&18.33&\textbf{13.14} &23.49 &36.25\\
SAM \& Gemma     &    & 1.711       & \textbf{2.872}       & \textbf{3.731}    & 4.251       & 6.588       & 9.545    & \textbf{5.309}       & \textbf{7.765 }      & \textbf{9.932 }   &16.11&\textbf{20.31}&\textbf{28.63}       \\
 \hline

 \rowcolor{gray!20}

VAE \& MDN-LSTM &        & 2.807       & 3.713       & {4.368 }     & 9.054       & 9.031      &{11.90}&26.58  &24.69&25.86&37.34 &37.42 &37.58\\
 \rowcolor{gray!20}

SAM \& MLP  &  & 2.366  & 3.435  & \textbf{4.321}  & 5.491  & 8.426  & 10.46  &8.344  & 13.26  & 23.39 & 20.69  & 28.30  & 37.27

\\
 \rowcolor{gray!20}

SAM \& LSTM  & $m$=4 & 2.835  & 4.876  & 6.700  & 5.545  & 9.418  & 14.42  & 9.004  & 14.11 & 24.51  & 21.98  & 30.71  & 51.05

\\
SAM \& GPT 3.5   &            & \textbf{1.683}       & 3.769       & 6.398       & \textbf{2.498}  &\textbf{6.307}    & 30.50    &\textbf{5.970 }&12.10&19.37&\textbf{10.99}&23.84  &39.76\\
SAM \& Gemma    &        & 1.813       & \textbf{2.954}       & 4.453       & 4.602       & 6.871       & \textbf{10.06}    & 6.153       & \textbf{7.818 }      & \textbf{12.34 }   &18.44&\textbf{22.30}&\textbf{31.00}  \\
 \hline

  \rowcolor{gray!20}

VAE \& MDN-LSTM &           & 3.682       & 3.584      & {4.893}       & 9.000       & 9.074    & 12.00 & 24.94 &25.36 &  25.01&37.30&37.47 &36.75 \\

 \rowcolor{gray!20}

SAM \& MLP  && 2.208&3.610&5.012&5.307&7.989&11.32& 19.87&11.77 &18.82 &14.73 &27.05& 36.12

\\
 \rowcolor{gray!20}

SAM \& LSTM  &$m$=8  & 2.485  & 5.008  & 7.701  &5.134  & 10.42  &16.16  &8.853  & 17.69  & 31.12  & 21.04  & 34.97  & 59.24

\\
SAM \& GPT 3.5  &             & 1.826      & 4.150       & 7.904       & \textbf{2.842 }      & \textbf{6.451}    &11.34&6.010  &13.08&16.55&\textbf{11.38 }&25.77 &35.57\\

SAM \& Gemma      &        & \textbf{1.787 }      & \textbf{3.034}       & \textbf{4.352}       & 4.798       & 7.153       & \textbf{10.32}    & \textbf{5.803 }      & \textbf{8.685 }      & \textbf{11.46}   &19.71&\textbf{24.92}&\textbf{30.16}     \\

        \bottomrule

\end{tabular}
\caption{State prediction performance for the cart pole: the mean centroid distance (CD) and the mean absolute error (MAE)  for the pole's angle. Gray rows indicate the use of additional data.}
\label{tab:physics}
\end{table*}

\looseness=-1
After inputting our prompt $\pi$, we will receive a formatted state prediction output $\hat{\bm{z}}$ based on which we evaluate the safety. To continue the prediction sequence into the next step, we need to generate an observation to feed into the image-based controller $h$. This ``output disassembly'' step is implemented with Algorithm~\ref{alg:disassemble}, and describes the process of rebuilding the observation for the controller. This kind of rebuilding won't cause object duplication and loss as shown in Fig~\ref{fig:shifted}, which is a common distribution shift that occurs in standard world models. This problem is brought about by the inability of the decoder to construct an observation for an out-of-distribution latent representation predicted by the surrogate model. Our proposed models use meaningful latent states, which are comparable based the CD metric, on which safety predicates can be encoded directly and evaluated without learning. This issue does not exist in the foundation world models because safety can be interpreted by the estimated states and the image reconstruction is precise pixel movement.


\begin{table*}[h]
\centering
\begin{tabular}{llllllll|llllll}
        \toprule

\multirow{3}{*}{} & \multirow{3}{*}{} & \multicolumn{6}{c}{F1 score $\uparrow$} & \multicolumn{6}{c}{FPR $\downarrow$}\\ 

        \midrule
          Method  &   Input length     & k=10     & k=20        & k=30     & k=40        & k=50     & k=60   & k=10     & k=20        & k=30     & k=40        & k=50     & k=60     \\ \hline
             \rowcolor{gray!20}

VAE \& MDN-LSTM   &  &.8610 & .8280 & .8159 & \textbf{.8099} & \textbf{.7912} & \textbf{.7963} & 1.000 & 1.000 & 1.000 & 1.000 & 1.000 & 1.000


\\
 \rowcolor{gray!20}

 SAM \& MLP &            & .0000 & .0000 & .0000 & .0000 & .0000 & .0000 & .3914 & .2845 & .2074 & .1002 & .0535 & \textbf{.0000} 
\\

 \rowcolor{gray!20}

  SAM \& LSTM &    $m$=1        & .0000 & .0000 & .0000 & .0000 & .0000 & .0000 & \textbf{.0676} & \textbf{.0784} & \textbf{.0579} & \textbf{.0579} & \textbf{.0289} & \textbf{.0000} \\
SAM \& GPT 3.5   &                &\textbf{.9566} & \textbf{.9324} & \textbf{.8622} & .7464 & .5328 & .6821 & .0863 & .2293 & .3000 & .2533 & .1006 & .2367  \\
SAM \& Gemma      &               & .9159 & .6667 & .7143 & .3670 & .5233 & .6471& 1.000 & 1.000 & 1.000 & .2381 & .7480 & .6561
\\
\hline
 \rowcolor{gray!20}

VAE \& MDN-LSTM   &  &
.8558 & .8345 & .8106 & .8067 & .7955 & .7771 & 1.000 & 1.000 & 1.000 & 1.000 & 1.000 & 1.000


\\
 \rowcolor{gray!20}

  SAM \& MLP &            & .0000 & .0000 & .0000 & .0000 & .0000 & .0000 & .4456 & .2692 & .2161 & .1320 & .0400 & \textbf{.0000} 
 \\
 \rowcolor{gray!20}

    SAM \& LSTM &      $m$=2      & .0000 & .0000 & .0000 & .0000 & .0000 & .0000 & .0685 & .0802 & .0536 & .0606 & \textbf{.0000} & \textbf{.0000} \\
SAM \& GPT 3.5   &                &  \textbf{.9280} & \textbf{.9094} & \textbf{.8371} & .8624 & .6479 & .6695 & \textbf{.0078} & .0443 & .0338 & .0248 & .0353 & .0206  \\
SAM \& Gemma           &               &    .8850 & .9038 & .3000 & \textbf{1.000} &\textbf{.8651} &\textbf{.8329}& .1441 & \textbf{.0000} & \textbf{.0000} & \textbf{.0000} &.0221 &.0375
\\

\hline
 \rowcolor{gray!20}

VAE \& MDN-LSTM   &  & .8535 & .8466 & .8013 & .7995 & .8088 & .7897 & 1.000 & 1.000 & 1.000 & 1.000 & 1.000 & 1.000


\\
 \rowcolor{gray!20}

  SAM \& MLP &            & .0000 & .0000 & .0000 & .0000 & .0000 & .0000 & .5473 & .4475 & .3860 & .3093 & .2871 & .2889  \\
   \rowcolor{gray!20}

    SAM \& LSTM &     $m$=4       & .0000 & .0000 & .0000 & .0000 & .0000 & .0000 & .0825 & .0837 & .0775 & .0597 & \textbf{.0000} & \textbf{.0000}\\
SAM \& GPT 3.5   &               & .9487 & .9471 & .7580 & \textbf{.8333} & .4537 & .5861 & \textbf{.0000} & .0075 & \textbf{.0000} & \textbf{.0061} & .0067 & \textbf{.0000}  \\
SAM \& Gemma           &               &\textbf{.9759} & \textbf{.9815} & \textbf{.9367} & .7929 & \textbf{.8843} & \textbf{.8137} & .1304 & \textbf{.0000} & .1750 & .6667 & .2889 & .6190 
 \\

\hline
 \rowcolor{gray!20}

VAE \& MDN-LSTM   &  &.8270 & .8113 & .8010 & .8045 & \textbf{.7923} & \textbf{.7797} & 1.000 & 1.000 & 1.000 & 1.000 & 1.000 & 1.000 


\\
 \rowcolor{gray!20}

 SAM \& MLP &            & .0000 & .0000 & .0000 & .0000 & .0000 & .0000 & \textbf{.0000} & \textbf{.0000} & \textbf{.0000} & \textbf{.0000} & \textbf{.0000} & \textbf{.0000} 
 \\
 \rowcolor{gray!20}
  SAM \& LSTM &    $m$=8        &.0000 & .0000 & .0000 & .0000 & .0000 & .0000 & .0527 & .0667 & .0742 & .0579 & .0460 & .0638  \\
SAM \& GPT 3.5   &                &  .9168 & \textbf{.9710} & .6575 & \textbf{.8233} & .3260 & .6042 & .0064 & .0068 & \textbf{.0000} & \textbf{.0000} & \textbf{.0000} & \textbf{.0000}  \\
SAM \& Gemma           &               &\textbf{.9677} & .7570 & \textbf{.9582} & .6047 & .5596 & .7009 & .1429 & .2391 & .1045 & .2016 & \textbf{.0000} & .2639  \\

        \bottomrule

\end{tabular}
\caption{Safety prediction performance for the lunar lander: F1 score and false positive rate. Gray rows indicate the use of additional data.}
\label{tab:lunarf1fpr}
\end{table*}

\section{Experiments and Results}\label{sec:exp}
\subsection{Experimental Setup}
We use simulation environments from OpenAI gym~\cite{DBLP:journals/corr/BrockmanCPSSTZ16} as two cases in this paper: (i) a cart pole with a four-dimensional physical state and (ii) a lunar lander with an eight-dimensional physical state. For the cart pole, the safety threshold $\theta_{thre}$ is defined as the angle between the pole and the vertical line $\theta$ being less than $\pi/4$: $|\theta|<\pi/4$. For the lunar lander, safety means keeping the horizontal position $d_x$ of the lander within the landing range: $8<d_x<12$. For the cart pole, to evaluate safety from our predicted states, we calculate the angle instead of showing the pole's centroid error. For the cart pole, given that the pole is upright most of the time, to get a more detailed evaluation of our models, we split the test experiment of the cart pole into two parts: (i) upright and (ii) falling.

As supervised learning baselines, we train two additional state prediction models: a Multilayer Perceptron (MLP) and a Long Short-Term Memory (LSTM). They are combined with our SAM-based representations. We also compare our approach to the original world model based on VAE representations and an MDN-LSTM state predictor. We test for 10,000 sequences with varied input lengths $m$ and prediction horizons $k$ in each case study. In the results tables, we marked the results of the supervised models in gray because these models take \textit{extra} 30,000 sequences to train and therefore impose a significant data burden compared to our zero-shot foundation world models. This fact relaxes the expectations of the performance of our proposed models.

\begin{table*}[h]
\centering
\begin{tabular}{lllll|lll|lll|lll}
        \toprule

\multirow{3}{*}{Method} & \multirow{3}{*}{Input length} & \multicolumn{6}{c}{F1 score $\uparrow$} & \multicolumn{6}{c}{FPR $\downarrow$}\\ & & \multicolumn{3}{c}{Upright} & \multicolumn{3}{c}{Falling }  & \multicolumn{3}{c}{Upright} & \multicolumn{3}{c}{Falling } \\

        \midrule
            &        & k=10     & k=20        & k=30     & k=10        & k=20     & k=30   & k=10     & k=20        & k=30     & k=10        & k=20     & k=30     \\ \hline
             \rowcolor{gray!20}

VAE \& MDN-LSTM   &  &.9987 & \textbf{.9915} & .9732 & .9905 & .9301 & .8598 & .0676 & .4474 & 1.000 & .0610 & .4756 & \textbf{.9980}


\\
 \rowcolor{gray!20}

 SAM \& MLP &            & .9964  & .9763  & .9317  &.9729  & .9113  & .5915  & .0000     & .\textbf{4328}  & 1.000  & .0000  & \textbf{.4167}  & 1.000 \\

 \rowcolor{gray!20}

  SAM \& LSTM &    $m$=1        & .9964  &.9789 & .9646 &.9729 &\textbf{.9382}  & .8636  & .0000  & .4328  & 1.000  & .0000  & .4167 & 1.000 \\
SAM \& GPT 3.5   &                &.9886  & {.9909}  &\textbf{ .9871} & .8837  & .8636  &\textbf{.8857}  & .7857& 1.000  &1.000 & .8407  & 1.000 & 1.000 \\
SAM \& Gemma      &               & \textbf{1.000} & .9531 & .9352  & \textbf{1.000 } & .6667  & .6021 & \textbf{.0000}  & 1.000  & 1.000 & \textbf{.0000}  & 1.000  & 1.000 \\
\hline
 \rowcolor{gray!20}

VAE \& MDN-LSTM   &  &
.9977 & .9912 & .9753 & \textbf{.9905} & .9318 & .8576 & .1125 & .4545 & 1.000 & .0610 & .4634 & 1.000 


\\
 \rowcolor{gray!20}

  SAM \& MLP &            & .9987  & \textbf{.9941}  & .8075  & {.9870} & \textbf{.9459}  & .6551  &.0704 & .0666 & \textbf{.0641} & .0833  & .0833  & .0833 \\
 \rowcolor{gray!20}

    SAM \& LSTM &      $m$=2      & \textbf{.9994}  & .9860  & .9670  &.9866  & .9295  & .8529  & \textbf{.0000 }& \textbf{.0000} & .0641 & \textbf{.0000} & \textbf{.0000}  & \textbf{.0833} \\
SAM \& GPT 3.5   &                & .9924 & .9768 & .9711  & .9398  & .8981 &.8662  & .4667  &.7333 & .7692 & .4182  & .7167  & .8333  \\
SAM \& Gemma           &               & .9926  &.9906  &\textbf{ .9873}  & .9473     & .9444  & \textbf{.9167 } &  .5625  &1.000  & 1.000  & .5000  & 1.000   & 1.000   \\

\hline
 \rowcolor{gray!20}

VAE \& MDN-LSTM   &  & \textbf{ .9992} & .9910 & .9778 & .9861 & .9324 & .8556 & \textbf{.0429} & .4730 & .9877 & .0854 & .4593 & .9980 


\\
 \rowcolor{gray!20}

  SAM \& MLP &            & {.9982} & \textbf{.9979} & \textbf{.9933}  & \textbf{.9870 } & \textbf{.9867 }& \textbf{.9743} & .0795  & \textbf{.0000}  & \textbf{.1941}  & \textbf{.0833}  & \textbf{.0000 }   & \textbf{.1667} \\
   \rowcolor{gray!20}

    SAM \& LSTM &     $m$=4       & .9982  &.9922  &.9346 & .9870  & .9382  & .5901  & .0795  & .4305  & .3980   & .0833 & .4167  & .4167 \\
SAM \& GPT 3.5   &               & .9918  & .9802  & .9625  & .9488  & .8946  & .8487  &.3500 & .7333  & .8182  & .3178  & .7593  & .8890  \\
SAM \& Gemma           &               & .9881 & .991  & .9765 & .9062  &.9230  & .8955  & .6440 & 1.000  & 1.000  & .7500  & 1.000 & 1.000 \\

\hline
 \rowcolor{gray!20}

VAE \& MDN-LSTM   &  &\textbf{.9992} & .9919 & .9773 & \textbf{.9895} & .9324 & .8579 & \textbf{.0484} & .3750 & 1.000 & \textbf{.0671} & .4593 & .9980 


\\
 \rowcolor{gray!20}

 SAM \& MLP &            & {.9987 } &\textbf{ .9979}  & \textbf{.9969} & {.9870} & \textbf{.9866}&  \textbf{.9743}  & .0625  &\textbf{ .0000} & .1428   & {.0833 } &  \textbf{.0000 } &  .1667 \\
 \rowcolor{gray!20}
  SAM \& LSTM &    $m$=8        & .9987  & .9908  & .9320  & .9870 & .9315  & .7619  & .0625  & .1052  &.\textbf{0833}  & .0833  & .0833  & \textbf{.0833 }\\
SAM \& GPT 3.5   &                & .9960  & .9744 & .9650 & .9516  & .8850 &  .8535  &.2500  & .8421 & 1.000 & .2375  & .8083  & .8958  \\
SAM \& Gemma           &               &.9871  & .9820  & .9829  & .9041 &.8823  & .9062  & .6617 &1.000  & 1.000  & .7000  & 1.000  &  1.000 \\

        \bottomrule

\end{tabular}
\caption{Safety prediction performance for the cart pole: F1 score and false positive rate. Gray rows indicate the use of additional data.}
\label{tab:f1fpr}
\end{table*}

\begin{figure}[tbh]
  \centering\includegraphics[width=0.9\linewidth]{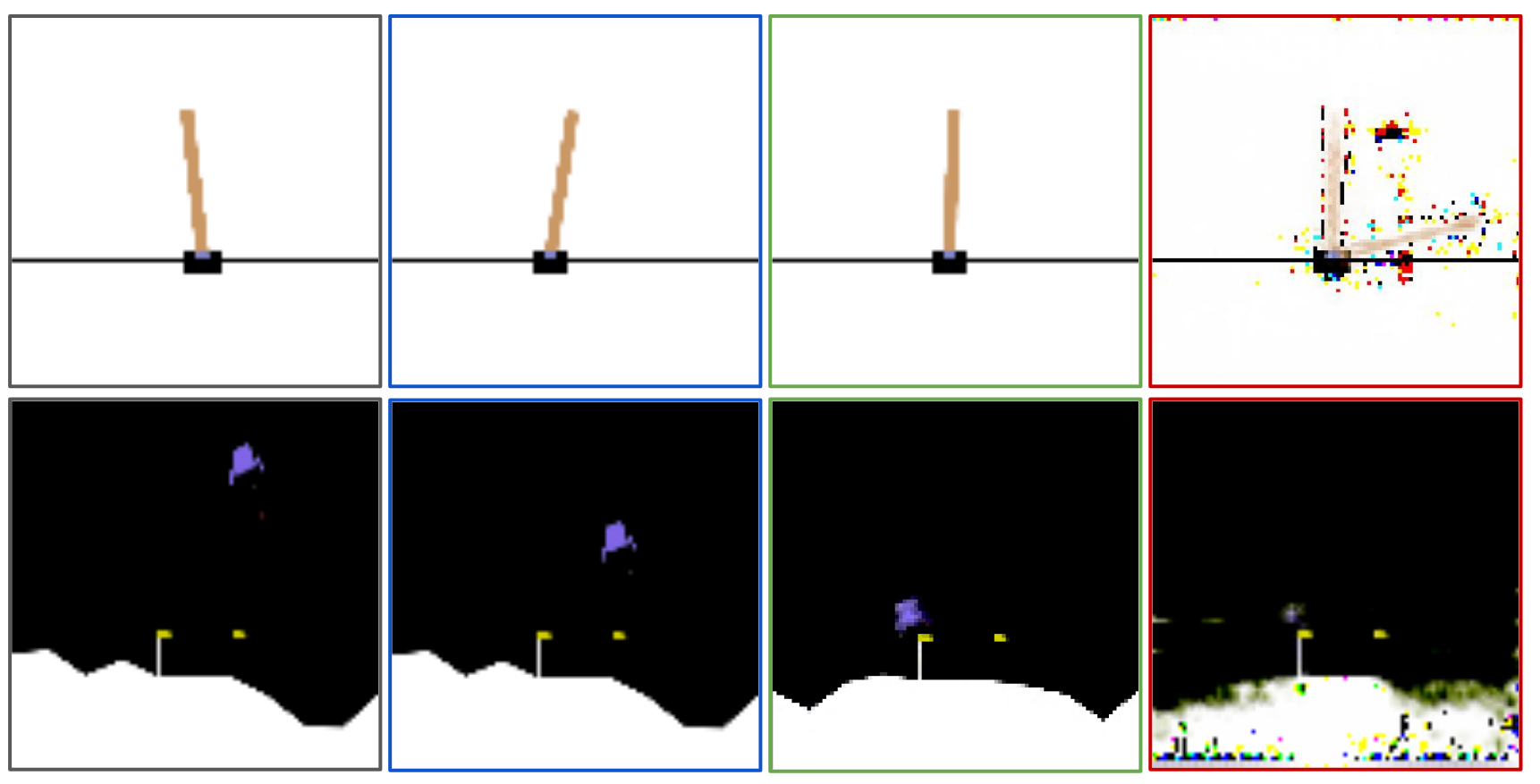}
  \caption{The observation of cart pole (upper) and lunar lander (lower). From left to right, generated by: the true observation model, a foundation world model, an existing world model without distribution shift, and an existing world model with distribution shift.}
    \label{fig:shifted}
\end{figure}

\subsection{Experimental Results}


First, we discuss the state prediction results. For the cart pole, in order to show the safety-related state more intuitively, we calculate the angle by the centroids of the cart and pole to substitute the CD of the pole. As depicted in Table~\ref{tab:cdlunar} and~\ref{tab:physics}, the errors of all methods are relatively low on short prediction horizons. As the value of $k$ increases, the error generally increases, indicating that predicting states further into the future is more difficult. Specifically, GPT 3.5 and Gemma alternately exhibit relatively low errors in most settings, especially when predicting the falling status.  Table~\ref{tab:cdlunar} shows the error for the lunar lander, where the foundation models do not perform as well as supervised learning methods on short inputs. Our investigation suggests that this shortcoming is due to insufficiently long input prompts, chosen due to cost constraints; accordingly, the Gemma-based approach beats the rest in most cases for input length 8. 

Tables~\ref{tab:lunarf1fpr} and~\ref{tab:f1fpr} show the results of the safety prediction. The standard world model performs well on short prediction horizons. As the horizon becomes longer, SAM-based models exhibit higher F1 and lower FPR. The foundation world models have competitive results compared with supervised learning despite being zero-shot. For the lunar lander, the standard world model and supervised learning fail to predict safety accurately. In the standard world model, the FPR is 1 for all predictions, which means that it predicts ``safe'' at all times. This might be caused by the vulnerability of generative ability~\cite{DBLP:journals/corr/abs-1812-08342} to distribution shifts~\cite{MORENOTORRES2012521} when the model encounters unseen data, with examples shown in Fig.~\ref{fig:shifted}. Another consideration is that, since the trajectory of the lunar lander resembles the shape of a sine function, this trajectory may exceed the safety bounds in intermediate states. The supervised learning methods they forecast a smoother trajectory without unsafe intermediate states, which leads to poor safety prediction because safety is defined on the whole sequence, but just on the last state.  

The secondary results with SSIM and MSE of predicted observations are shown in Tables~\ref{tab:lunarSSIM} and~\ref{tab:SSIM1}. As mentioned earlier, these metrics are not relevant enough to the dynamics and safety predictions of safety-critical systems. Combining with the metric of the state prediction, it shows that he standard world model can only do well in the MSE and fails in SSIM, which means it has a low quality of predicting and reconstructing the observations.

\begin{table*}[th!]
\centering
\begin{tabular}{llllllll|llllll}
        \toprule

\multirow{3}{*}{} & \multirow{3}{*}{} & \multicolumn{6}{c}{SSIM $\uparrow$} & \multicolumn{6}{c}{MSE $\downarrow$}\\   

        \midrule
        Method   &     Input length   & k=10     & k=20        & k=30     & k=40        & k=50     & k=60   & k=10     & k=20        & k=30     & k=40        & k=50     & k=60     \\ \hline
             \rowcolor{gray!20}

 VAE \& MDN-LSTM  & & .6933 & .6933 & .6931 & .6933 & .6932 & .6933 & .1925 & .1924 & .1925 & .1925 & .1924 & .1924

\\
 \rowcolor{gray!20}

SAM \& MLP  &  & .9922 & .9901 & .9896 & .9892 & .9890 & .9888 & .0013 & .0021 & .0023 & .0024 & .0025 & .0025

\\
 \rowcolor{gray!20}

SAM \& LSTM  &  $m$=1 &.9930 & \textbf{.9914} & \textbf{.9905} & .9900 & .9897 & \textbf{.9894} & .0010 & \textbf{.0016} & \textbf{.0020} & \textbf{.0021} & .0022 & \textbf{.0023}

\\
SAM \& GPT 3.5   &              &   .9923 & .9911 & \textbf{.9905} &\textbf{.9903} & \textbf{.9909} & .9891 & .0015 & .0019 & .0022 & .0022 & \textbf{.0020} & .0026 \\
SAM \& Gemma      &               &  \textbf{.9950} & .9892 & .9896 & .9893 & .9902 & .9889 & \textbf{.0007} & .0024 & .0027 & .0024 & .0022 & .0026 
 \\
\hline
 \rowcolor{gray!20}

VAE \& MDN-LSTM   && .6933 & .6933 & .6932 & .6933 & .6932 & .6932 & .1924 & .1924 & .1924 & .1925 & .1925 & .1924 

            
            \\
 \rowcolor{gray!20}

SAM \& MLP  &  & 
.9929 & .9911 & .9902 & .9897 & .9894 & .9891 & .0010 & .0017 & .0021 & .0022 & .0024 & .0025 
\\

 \rowcolor{gray!20}

SAM \& LSTM  & $m$=2 &  .9924 & .9907 & .9899 & .9896 & .9894 & .9891 & .0012 & .0018 & .0021 & .0022 & .0023 & .0024

\\
SAM \& GPT 3.5   &               & \textbf{.9947} & \textbf{.9926} & \textbf{.9915} & \textbf{.9911} & \textbf{.9906} & \textbf{.9908} & \textbf{.0008} & \textbf{.0015} & \textbf{.0018} & \textbf{.0020} & \textbf{.0022} & \textbf{.0022}   \\
SAM \& Gemma           &               & .9918 & .9907 & .9897 & .9884 & .9891 &.9834 & .0017 & .0021 & .0021 & .0028 & .0025 & .0027
 \\

\hline
 \rowcolor{gray!20}

VAE \& MDN-LSTM   & & .6932 & .6933 & .6932 & .6933 & .6933 & .6933 & .1925 & .1925 & .1924 & .1925 & .1925 & .1925


\\
 \rowcolor{gray!20}

SAM \& MLP  &  &
.9936 & .9921 & \textbf{.9910} & .9904 & .9901 & \textbf{.9898} & .0008 & .0014 & \textbf{.0018} & \textbf{.0020} & \textbf{.0021} & .0023 
\\
 \rowcolor{gray!20}

SAM \& LSTM  & $m$=4 & .9927 & .9913 & .9904 & .9900 & .9896 & .9894 & .0011 & .0016 & .0020 & .0021 & .0022 & .0023 

\\

SAM \& GPT 3.5   &                &  \textbf{.9954} & \textbf{.9932} & .9908 & .9884 & .9900 & .9862 & \textbf{.0006} & \textbf{.0012} & .0020 & .0021 & .0023 & \textbf{.0021} 
\\
SAM \& Gemma           &               & .9920 & .9930 & .9909 & \textbf{.9908} & \textbf{.9905} & .9890 & .0016 & .0013 & .0020 & \textbf{.0020} & \textbf{.0021} & .0026  \\

\hline
 \rowcolor{gray!20}

VAE \& MDN-LSTM   & & .6932 & .6933 & .6932 & .6932 & .6932 & .6932 & .1925 & .1924 & .1924 & .1924 & .1925 & .1925


\\
 \rowcolor{gray!20}

SAM \& MLP  &  & \textbf{.9936} & \textbf{.9940} & \textbf{.9940} & \textbf{.9940} & \textbf{.9940} & \textbf{.9940} & .0015 & .0014 & \textbf{.0014} & \textbf{.0014} & \textbf{.0014} & \textbf{.0014} 

\\
 \rowcolor{gray!20}

SAM \& LSTM  & $m$=8 & .9929 & .9914 & .9906 & .9900 & .9894 & .9890 & .0011 & .0015 & .0019 & .0021 & .0023 & .0025

\\
SAM \& GPT 3.5   &               & .9847 & .9847 & .9773 & .9565 & .9623 & .9101 & \textbf{.0004} & \textbf{.0012} & .0022 & .0021 & .0024 & .0020  \\
SAM \& Gemma         &               &.9908 & .9907 & .9900 & .9891 & .9890 & .9894 & .0019 & .0021 & .0023 & .0025 & .0025 & .0024 
 \\

        \bottomrule

\end{tabular}
\caption{Comparison of Structural similarity index measure (SSIM) and mean square error (MSE) for lunar lander's predicted observations. Gray rows indicate the use of additional data.}
\label{tab:lunarSSIM}
\end{table*}

\section{Related Work}\label{sec:realated}

\subsection{Foundation Models and Applications}


With the rise of large AI models that benefit from the availability of huge computing resources by using a considerable amount of data and model parameters, foundation models achieve high performance on multiple cross-domain tasks. Many well-performing models like Generative Pre-trained Transformer (GPT)~\cite{brown2020language}, Bidirectional Encoder Representations from Transformers (BERT)~\cite{devlin2019bert}, Contrastive Language-Image Pre-Training  (CLIP)~\cite{DBLP:journals/corr/abs-2103-00020} and Segment Anything model (SAM)~\cite{kirillov2023segment} extensively serve as the \textit{foundation} for downstream tasks. 
We extend the core idea of LLMs from text prediction to complex sequence tasks such as safety prediction, by exploiting the statistical patterns of language learned from large text data.

\looseness=-1
Furthermore, one significant advantage of LLMs is their extensive pre-training, making them easily adapted to various domains such as  
robotics decision-making tasks~\cite{mirchandani2023large}. Another language-guided abstraction~\cite{peng2024learning} can transfer high-level task descriptions into task-relevant state abstractions by using a pre-trained language model.
Abstract meaning representation also plays an important role in LLMs and can improve their performance~\cite{jinrole}. 
We believe that LLNs can achieve better performance on complex task sequences because the alignment between human and robot representations is closer than that of human and robot understanding~\cite{Bobu_2024}.

As the internal structure of LLM is a generative pre-trained transformer, it can be used for any prediction probelms. Some researchers have tested LLM's forecasting ability with humans and it shows that the ability of LLM is approaching the human level~\cite {halawi2024approaching}. Some new foundation models like Chronos~\cite{ansari2024chronos} are specialized in non-dynamical prediction tasks like traffic, weather, and energy consumption. Language can also be the input of robotic systems~\cite{vemprala2023chatgpt} to be trained for end-to-end language-controlled systems like RT2~\cite{brohan2023rt2}, which can perform basic semantic reasoning to finish its tasks. The trustworthiness of robotic techniques based on foundation models is still an area that has not been widely explored~\cite{huang2023survey}. Besides applying the foundation model to control and forecast,  novel simulators also adopt LLMs to generate various scenes for traffic simulations~\cite{tan2023language}.

\subsection{Safety Assurance of Autonomous Systems}

Trajectory forecasting plays a key role within autonomous systems for a variety of purposes, including safety assurance. One tricky problem with safety assurance based on trajectory prediction is the gap between high-dimensional input and low-dimensional states.
For sensors operating in high-dimensional spaces, integrating physics models is crucial for enhancing predictive accuracy. Works such as the Social ODE~\cite{wen2022social} and Deep Kinematic Models~\cite{9197560} exemplify this approach by merging deep learning techniques with physical models to address issues related to data unreliability. 
Another popular safety assurance method is reachability analysis. Given the challenge of finding all reachable states in non-linear dynamics and neural network controllers, specialized verification tools~\cite{althoff, flow} are designed to develop accurate overapproximations of these reachable sets. However, these tools cannot work on high-dimensional controllers such as image-input controllers. One tricky problem with safety assurance based on trajectory prediction is the gap between high-dimensional input and low-dimensional states.
High-dimensional verification processes could be tailored, for instance, through the application of generative models~\cite{DBLP:journals/corr/abs-2105-07091}, and by approximating high-dimensional systems with their lower-dimensional counterparts~\cite{geng2024bridging}. 
Unlike the cases above, we apply SAM to bridge the gap between high-dimensional and low-dimensional space, which requires no additional training like LLNs and is easy to implement.

Finally, conformal prediction is also widely recognized for establishing boundaries around time series forecasts within a defined confidence level ~\cite{dixit2023adaptive, qinxin}, which could be used to provide safety bounds for the trajectory with a specific confidence.

\begin{table*}[th!]
\centering
\begin{tabular}{lllll|lll|lll|lll}
        \toprule

\multirow{3}{*}{Method} & \multirow{3}{*}{Input length} & \multicolumn{6}{c}{SSIM $\uparrow$} & \multicolumn{6}{c}{MSE $\downarrow$}\\ & & \multicolumn{3}{c}{Upright} & \multicolumn{3}{c}{Falling }  & \multicolumn{3}{c}{Upright} & \multicolumn{3}{c}{Falling } \\

        \midrule
            &        & k=10     & k=20        & k=30     & k=10        & k=20     & k=30   & k=10     & k=20        & k=30     & k=10        & k=20     & k=30     \\ \hline
             \rowcolor{gray!20}

 VAE \& MDN-LSTM  & & .8678 & .8629 & .8567 & .8426 & .8332 & .8297 & .0031 & .0049 & .0058 & .0060 & .0094 & .0103

\\
 \rowcolor{gray!20}

SAM \& MLP  &  & \textbf{.9779}  & \textbf{.9612 } & .9494  & \textbf{.9526}  & .9374  & .9300  & .0023  & .0045  & .0060 & \textbf{.0050}  & .0074  & .0086

\\
 \rowcolor{gray!20}

SAM \& LSTM  &  $m$=1 &.9757  & .9575 & .9451 & .9487  & \textbf{.9411}  & \textbf{.9360}  & \textbf{.0022} &\textbf{.0044}  & .0061  & .0051 & \textbf{.0061}  & \textbf{.0070}

\\
SAM \& GPT 3.5   &              & .9691  & .9611  & \textbf{.9564} & .9314  & .9225  & .9234  & .0034 & .0048  &.\textbf{0055 } & .0078  & .0099  & .0098  \\
SAM \& Gemma      &               & .9632  & .9465  &.9322 & .9409  & .9140  & .9076  & .0035 & .0059  & .0066 &.0067  & .0096  & .0121  \\
\hline
 \rowcolor{gray!20}

VAE \& MDN-LSTM   && .8683 & .8629 & .8564 & .8437 & .8327 & .8294 & .0030 & .0049 & .0059 & .0059 & .0094 & .0103 

            
            \\
 \rowcolor{gray!20}

SAM \& MLP  &  & .9567  &.9316  & .9116  & .9337  & .9208 & .9045  & .0045  & .0069  & .0092  & .0065     & \textbf{.0076 } & .0100

\\

 \rowcolor{gray!20}

SAM \& LSTM  & $m$=2 & .9624  & .9403  & .9170  & .9333  & .9280  & .9137  & .0042  & .0068  &.0089  & .0074  & .0081 & \textbf{.0089}

\\
SAM \& GPT 3.5   &               & .9600 & .9383  & .9248  & .9400  & .9194  & .9045  & .0038  & .0064 & .0083 &  \textbf{.0057 } & .0086  & .0115  \\
SAM \& Gemma           &               & \textbf{.9724}  &\textbf{ .9623}  & \textbf{.9575}  & \textbf{.9418}  & \textbf{.9335}  & \textbf{.9231 } & \textbf{.0030}  & \textbf{.0046}  & \textbf{.0053}  & .0066  & .0086  & .0099  \\

\hline
 \rowcolor{gray!20}

VAE \& MDN-LSTM   & & .8682 & .8624 & .8574 & .8432 & .8331 & .8278 & \textbf{.0031} & .0049 & \textbf{.0057} & .0059 & .0094 & .0103


\\
 \rowcolor{gray!20}

SAM \& MLP  &  &.9596 & .9369  & .9177  & .9292  & .9177  & .9128  & .0042 & .0064 & .0076  & .0079  & .0100  &\textbf{.0101}

\\
 \rowcolor{gray!20}

SAM \& LSTM  & $m$=4 & .9539  & .9281 & .9160  & .9207  & .9020  & .8944  &.0050 & .0079  & .0091  &.0086 &.0116 & .0122

\\

SAM \& GPT 3.5   &                & .9594  & .9366  & .9200  & \textbf{.9424 } & .9157  & .9020  & .0039  & .0066  & .0087  & \textbf{.0054 } &  .0091  & .0115  \\
SAM \& Gemma           &               & \textbf{.9712}  & \textbf{.9614}  & \textbf{.9524 } &.9380  &\textbf{.9294}  & \textbf{.9213}  & {.0032}  & \textbf{.0047}  & .0060  & .0070 & \textbf{.0089}  & .0102  \\

\hline
 \rowcolor{gray!20}

VAE \& MDN-LSTM   & & .8679 & .8615 & .8566 & .8435 & .8320 & .8288 & \textbf{.0031} & .0049 & \textbf{.0057} & \textbf{.0058} & .0093 & \textbf{.0104}


\\
 \rowcolor{gray!20}

SAM \& MLP  &  & .9635 & .9441  & 9266 & .9314 & .9214  & {.9139} & .0039  &.0061 & .0078 & .0078  & .0098  & .0108

\\
 \rowcolor{gray!20}

SAM \& LSTM  & $m$=8 & .9553 & .9295  & .9146  & .9249  & .9006 & .8957  & .0047  & .0079  & .0095  & .0083 & .0122  & .0125

\\
SAM \& GPT 3.5   &               & .9577  & .9354  & .9220  & .9379  & .9140 &  .9002  & .0041  & .0068  & .0084 & {.0059}  & .0092  & .0114  \\
SAM \& Gemma         &               & \textbf{.9712}&\textbf{.9611} &\textbf{.9522} &\textbf{.9342 } &\textbf{ .9288}  & \textbf{.9170 }  &{.0032}  &\textbf{.0048} & .0061   &.0073  & \textbf{.0091}  & .0109 \\

        \bottomrule

\end{tabular}
\caption{Comparison of Structural similarity index measure (SSIM) and mean square error (MSE) for cart pole's predicted observations. Gray rows indicate the use of additional data..}
\label{tab:SSIM1}
\end{table*}

\section{Discussion}\label{sec:discussion}

    While our foundation world model demonstrates improved performance in certain aspects and produces physically meaningful states, its dynamic predictions essentially rely on statistical correlations. Also, deploying foundation world models may require human judgment about the relevance of segmented classes; however, for many robotic systems, this effort is dwarfed by the significant data collection required to deploy standard world models or other supervised methods.

    \looseness=-1
    In our future work, we plan to delve deeper into the fine-tuning of foundation models to enhance their performance on particular robotic prediction tasks. Additional dynamic states, such as speed and angular velocity, can be derived from the observation sequence and integrated into state representation to develop a more physics-specific surrogate model. In addition, other kinds of multimodal foundation models ~\cite{DBLP:journals/corr/abs-2103-00020,ramesh2021zeroshot} may also help with building an comprehensive world model.

\bibliographystyle{IEEEtran}

\bibliography{sample}

\end{document}